\documentclass[final]{cvpr}

\usepackage{times}
\usepackage{epsfig}
\usepackage{graphicx}
\usepackage{amsmath}
\usepackage{amssymb}
\usepackage{booktabs}
\usepackage{multirow}
\usepackage{color}
\usepackage{array}
\usepackage[ruled]{algorithm2e}
\usepackage{url}
\usepackage{stfloats}

\usepackage[pagebackref=true,breaklinks=true,colorlinks,bookmarks=false]{hyperref}

\begin{document}

%%%%%%%%% TITLE
\title{MLFW: A Database for Face Recognition on Masked Faces}
\author{Chengrui~Wang,
Han~Fang,
Yaoyao~Zhong,
Weihong~Deng\\
Beijing University of Posts and Telecommunications\\
{\tt\small \{crwang, fanghan, zhongyaoyao, whdeng\}@bupt.edu.cn}}

\maketitle
% Face recognition has been widely used in identity authentication to prevent intrusion.
% However, most of the current face recognition systems are designed for unmasked faces, and may encounter severe performance degradation when recognizing masked faces.
%  by searching images of identities in LFW with apparent age gaps to form positive pairs and selecting negative pairs using individuals with the same race and gender.

%%%%%%%%% ABSTRACT
\begin{abstract}
	% Cross-Age LFW (CALFW) database has been widely utilized as a benchmark of unconstrained face verification. 
	As more and more people begin to wear masks due to current COVID-19 pandemic, existing face recognition systems may encounter severe performance degradation when recognizing masked faces. To figure out the impact of masks on face recognition model, we build a simple but effective tool to generate masked faces from unmasked faces automatically, and construct a new database called Masked LFW (MLFW) based on Cross-Age LFW (CALFW) database.  
	The mask on the masked face generated by our method has good visual consistency with the original face. Moreover, we collect various mask templates, covering most of the common styles appeared in the daily life, to achieve diverse generation effects. Considering realistic scenarios, we design three kinds of combinations of face pairs. The recognition accuracy of SOTA models declines 5\%-16\% on MLFW database compared with the accuracy on the original images. MLFW database can be viewed and downloaded at \url{http://whdeng.cn/mlfw}.
\end{abstract}

%%%%%%%%% BODY TEXT
\section{Introduction}

COVID-19 pandemic forces people to start wearing masks to prevent themselves from the disease.
However, a mask will occlude a part region of a face, and change the facial features that can be obtained by face recognition models.
This raises concerns about whether a face recognition model can work well on masked faces~\cite{zhu2021masked,deng2021masked,boutros2021mfr}, especially when the model has never seen any masks during the training process.
In order to maintain the security of face recognition, the most important method in the contactless authentication, it is urgent to come up with a credible approach to evaluate the performance of face recognition models on the masked faces.

\begin{figure}
	\begin{center}
		\includegraphics[width=1\linewidth]{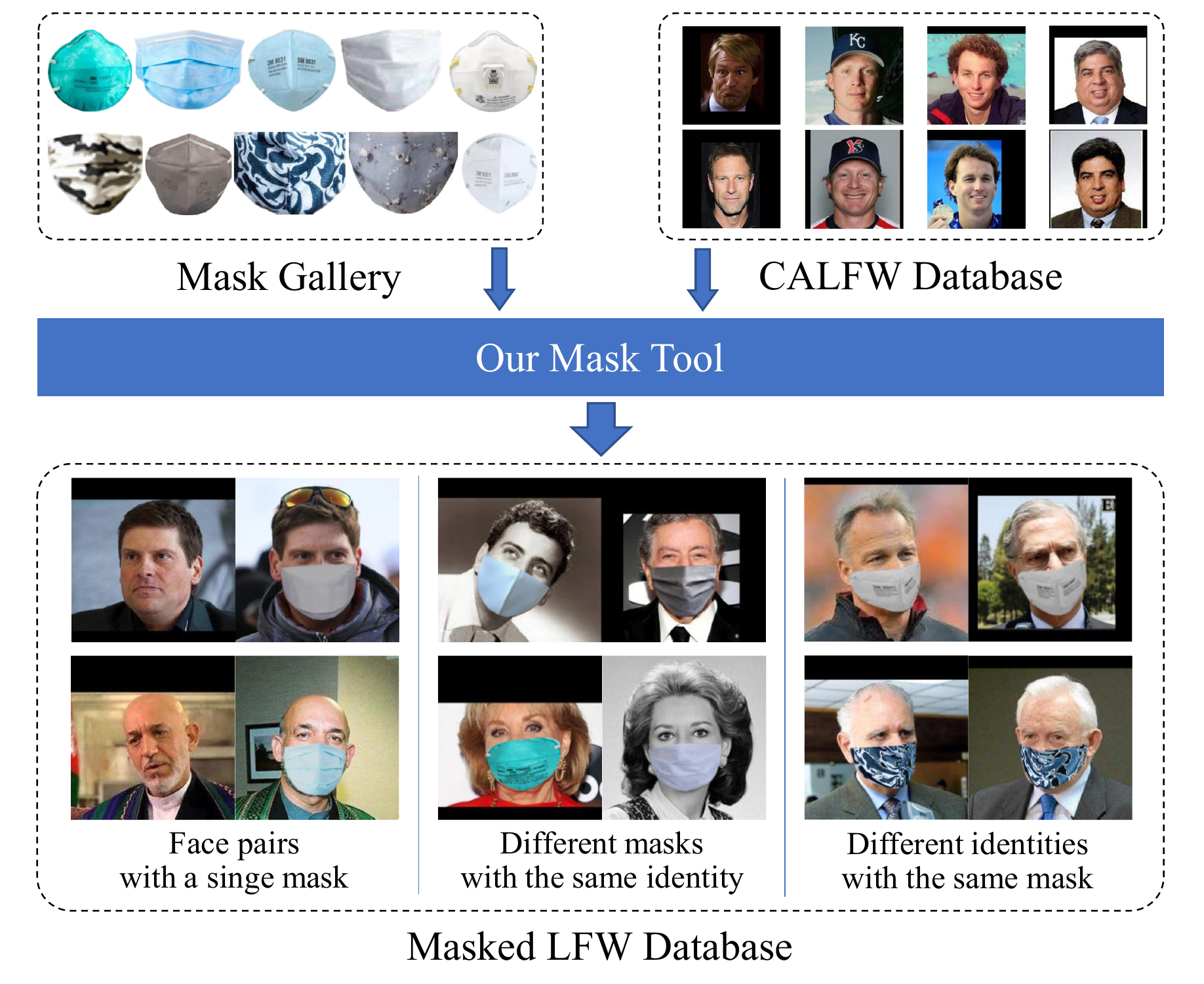}
	\end{center}
	\caption{
		With the mask tool, we remake the face images in CALFW database with masks in a gallery, and construct a new database, Masked LFW (MLFW) database, which shares the face verification protocol and the same identities in LFW database.
	}
	% \vspace{-0.5cm}
	\label{illustration}
\end{figure}

Recently, lots of works focus on generating masked face images and constructing related datasets.
Anwar~\etal~\cite{anwar2020masked} present an opensource tool to mask faces and create a large dataset of masked faces.
Wang~\etal~\cite{wang2020masked} propose three types of masked face datasets.
Du~\etal~\cite{du2021towards} use 3D face reconstruction to add the mask templates on the UV texture map of the non-masked face.
Cabani~\etal~\cite{cabani2021maskedface} provide a fine-grained dataset for detecting people wearing masks and those not wearing masks.
However, existing tools usually transform the whole mask to generate masked faces, resulting in unrealistic generation effect to a certain extent. In addition, there lacks a specialized dataset to verify the performance of face recognition models.

Face identification and face verification are the basic paradigms in face recognition.
Face identification accesses the identity of a query image through finding the most similar face in a gallery, while face verification attempts to verify whether two given face images have the same identity.
The protocol of face verification makes it easy to construct a persuasive database used for verifying the performance of face recognition models on masked faces.
Labeled Faces in the Wild (LFW) database~\cite{huang2008labeled} is a widely-used benchmark for face verification, which evaluates the performance of face recognition models in the unconstrained scenarios, and has attracted a lot of researchers.
Beyond that, several databases have been proposed based on LFW for special scenarios~\cite{deng2017fine,zheng2017cross,zheng2018cross,zhong2020towards}.
Particularly, Cross-Age LFW \cite{zheng2017cross} (CALFW) database adds age intra-class variation into the positive pairs of the original LFW database, which better simulates the situation of face verification in the real world.

In this paper, we reinvent the CALFW database and construct a new database called Masked CALFW (MLFW) through adding masks to some selected images of CALFW database.
The illustration of MLFW database is shown in Figure~\ref{illustration}.
Specifically, we consider three verification scenarios, 1) two faces have the same identity but wear different masks, 2) two faces have different identities but wear the same mask, 3) one face has a mask but the other does not.
For the 6,000 face pairs in CALFW database, we randomly divide them into three subsets, which contain 3,000, 1,500 and 1,500 face pairs respectively according to different scenarios.
The division reflects the hard examples in the real world and increases the difficulty of face verification.

\begin{figure*}[t]
	\begin{center}
		\includegraphics[width=1\linewidth]{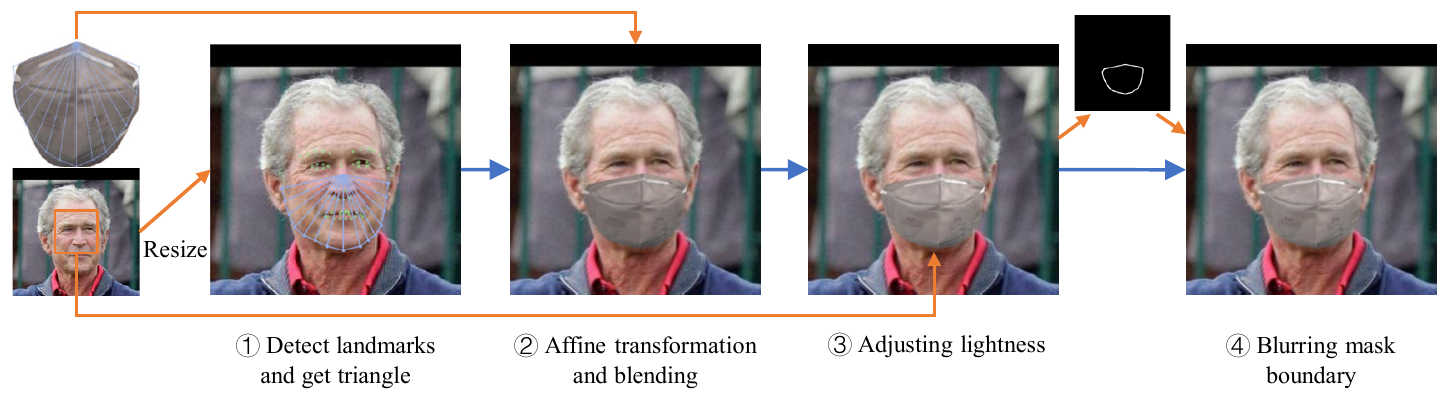}
	\end{center}
	\caption{
		The main procedure of our mask tool.
	}
	% \vspace{-0.4cm}
	\label{algorithmimg}
\end{figure*}

To construct the database, we build a tool to automatically generate masked faces from unmasked faces.
In the process of wearing mask, the tool divides both mask and face into multiple triangles separately according to landmarks, and adds each patch of mask to the correlative facial patch through affine transformation, which produces a appropriate fit between mask and face.
Besides, the tool adjusts the brightness of mask and smooths the edge between mask and face to make mask look more consistent with face.
For achieving diverse generation effects, we construct a mask gallery which contains various pre-processed masks and propose a method to interfere with the mask position.

The main contributions of this paper are as follows:

- We provide a tool to automatically generate masked face from unmasked face, which achieves real and diverse generation effects.

- Based on the proposed tool, we build a masked version called MLFW to evaluate the performance of face recognition model on masked faces.

\begin{figure*}[t]
	\begin{center}
		\includegraphics[width=1\linewidth]{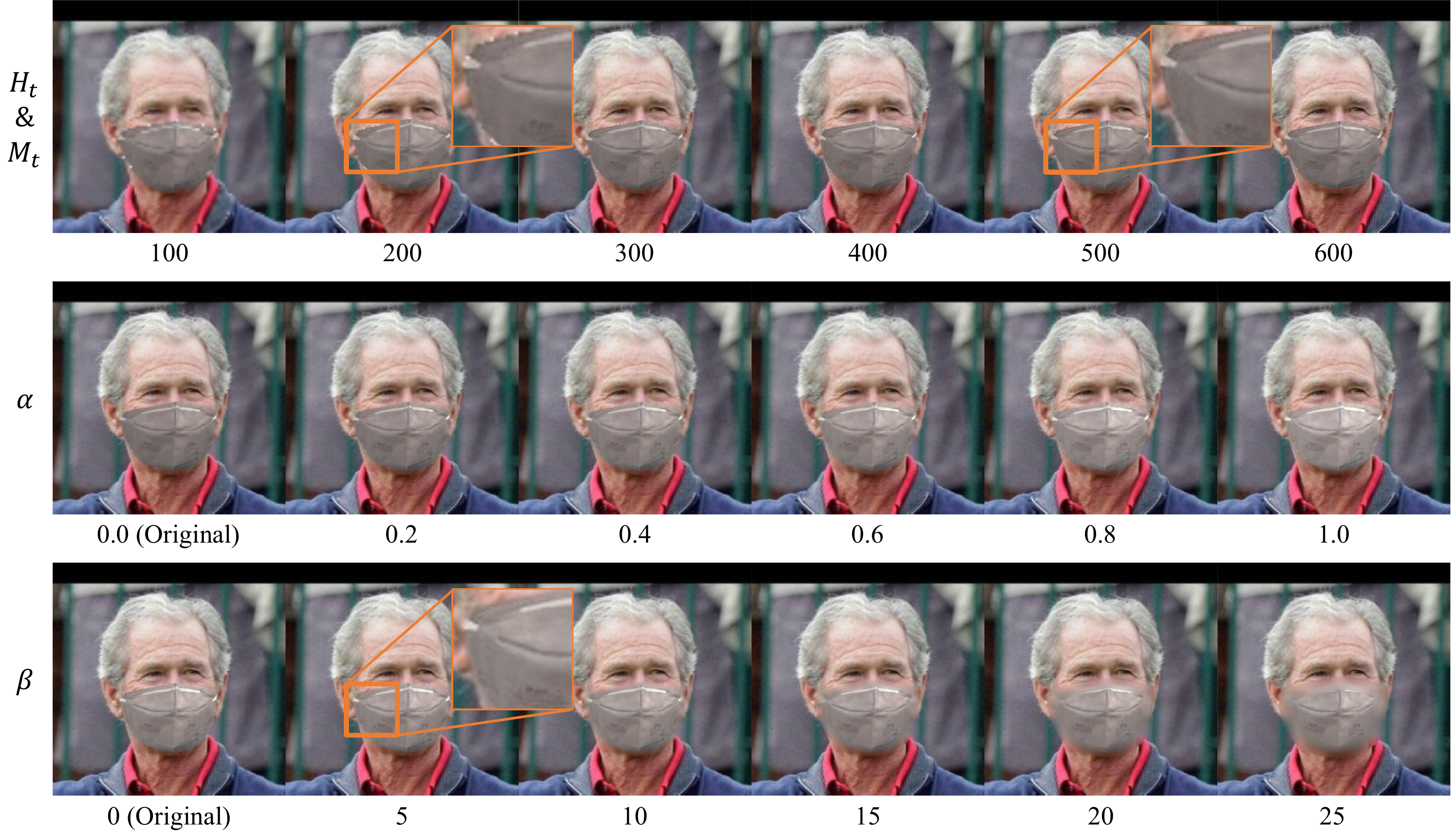}
	\end{center}
	\caption{
		Mask faces generated under different parameters $H_{t} \& W_{t}$ (top), $\alpha$ (center) and $\beta$ (bottom).
	}
	\label{params}
\end{figure*}

\begin{figure}[t]
	\begin{center}
		\includegraphics[width=0.9\linewidth]{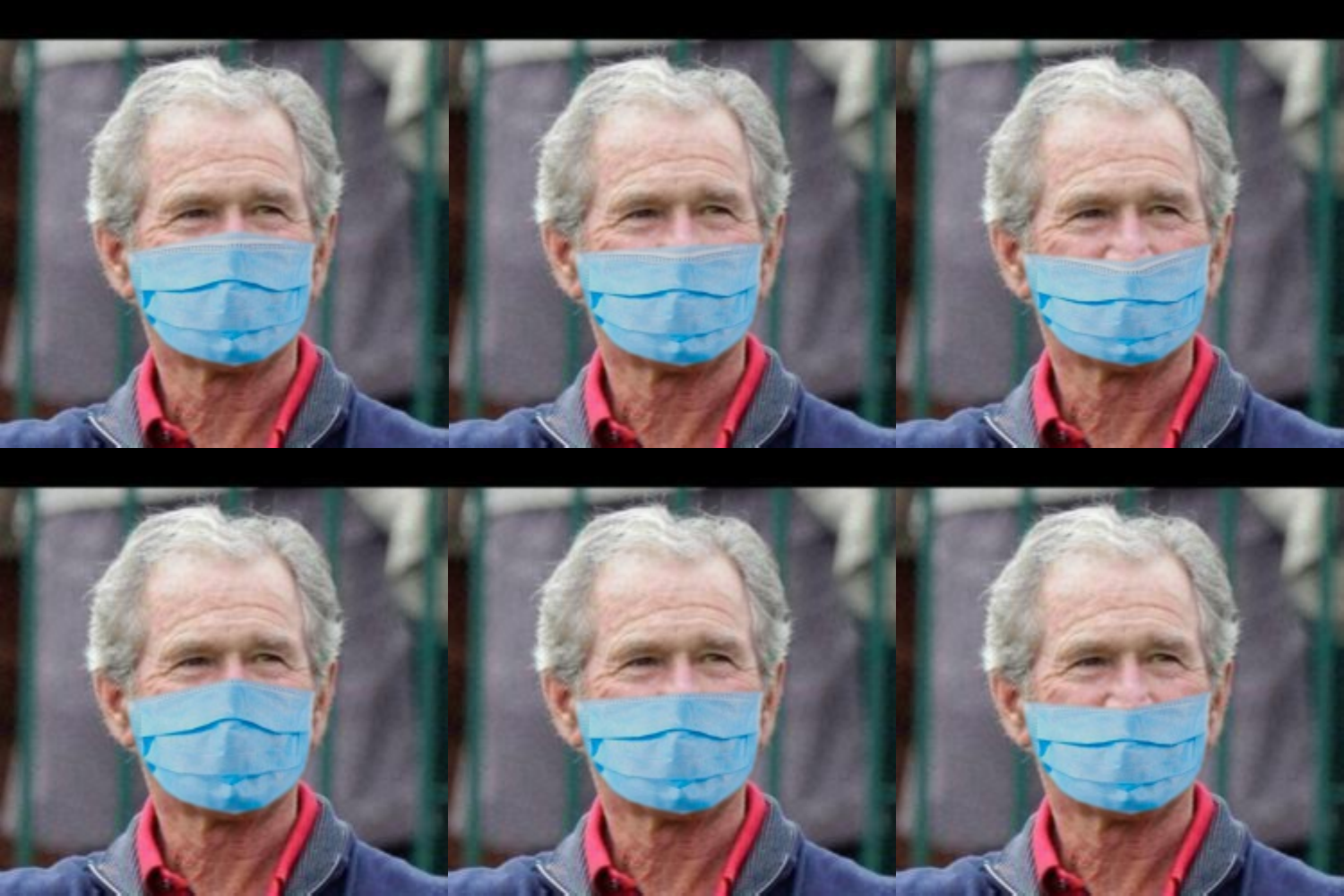}
	\end{center}
	\caption{
		The example mask faces generated under landmark perturbation.
	}
	\label{varietyexamples}
\end{figure}

\section{Mask Tool}
\subsection{Main procedure}
In this section, we detail the main procedure of our mask tool. The brief illustration is shown in Figure~\ref{algorithmimg}, which consists of four steps.

\textbf{Detect landmarks and get triangles.} Given input face image $I_f$, we firstly resize $I_f$ to $I_{f}^{'}$ of larger size $H_{t} \times W_{t}$, which aims to reduce aliasing in the output masked face, and detect landmarks of $I_f^{'}$ through a 68-points detector~\cite{bulat2017far}.
Then, according to specific landmarks set on $I_m$, we extract 14 patches from both $I_{f}^{'}$ and $I_m$, and calculate a transformation matrix for each pair of patches.

\textbf{Affine transformation and blending.} After that, we project the mask $I_m$ to $I_f^{'}$ to generate the masked face image $I_o^{'}$ based on the affine transformation.

\textbf{Adjusting lightness.} Next, we calculate the average value of L-channel on the center facial region of $I_f$ to adjust the L-channel of mask in $I_o^{'}$ under the control of weight $\alpha$.

\textbf{Blurring mask boundary.} Furthermore, in order to further reduce the visual inconsistency, we apply Gaussian Blur to blur the boundary between mask and face in $I_o^{'}$ with kernel size $\beta$.
Finally, we resize $I_o^{'}$ to the output size.

The visual difference of masked face generated under different parameters are shown in Figure~\ref{params}. In this work, $H_{t} \& W_{t}, \alpha$ and $\beta$ are set to $500$, $0.6$ and $5$, respectively.

\subsection{Generation variety}
We achieve various generation effect through 1) perturbing the landmark at the top of face, 2) perturbing the landmark at the top of mask, and 3) using different mask templates.
Some examples are shown in Figure~\ref{varietyexamples}.

\section{MLFW database}

With the help of the mask tool, we construct MLFW database based on CALFW database, and the main process can be described into the following steps.

\textbf{Dividing faces into three subsets.}

CALFW database contains 6,000 face pairs (3,000 positive pairs and 3,000 negative pairs).
We spilt the 6,000 pairs into three subsets, which contain 3,000, 1,500, 1,500 face pairs separately, to construct our MLFW database.

The first subset of MLFW database is designed to test whether a face recognition model can verify the identities of two faces when one of the faces wears a mask.
Therefore, we only add a mask to one face image for each face pair.
Specially, in this subset, the number of positive pairs is the same as negative pairs (1,500 positive pairs and 1,500 negative pairs).

The second and third subsets of MLFW database are designed to evaluate the performance of face recognition models in extremely hard cases that two faces with the same identity wear different masks and two faces with different identities wear the same mask, respectively.
Compared with adding masks randomly, experimental results show that the accuracy of SOTA models is further reduced by at least 2\% with our elaborate strategy.

Overall, the statistic of our MLFW database is shown in Table~\ref{tab:table0}. 
In addition, our MLFW dataset has been equally divided into 10 separate folds for cross validation.

\begin{table}[bh]
	\begin{center}
		\begin{tabular}{c|cc}
			\toprule
			Mask count & Positive Pair & Negative Pair\\
			\midrule
			1 & 1500 & 1500 \\
			2 & 1500 & 1500 \\\bottomrule
		\end{tabular}
	\end{center}
	\caption{Statistic of our MLFW database. \textit{Mask count} represents the count of masked faces in each face pair. }
	\label{tab:table0}
\end{table}

\begin{table*}[t]
	\begin{center}
\setlength{\tabcolsep}{1.8mm}{
\begin{tabular}{@{}l|cccccc@{}}
\toprule
                                          & LFW~\cite{huang2008labeled}              & SLLFW~\cite{deng2017fine}            & TALFW~\cite{zhong2020towards}            & CPLFW~\cite{zheng2018cross}            & CALFW~\cite{zheng2017cross}            & MLFW            \\ \midrule
{Private-Asia, R50, ArcFace}~\cite{wang2021face}        & {99.50} & {98.00} & {69.97} & {84.12} & {91.12} & {74.85} \\
{CASIA-WebFace, R50, CosFace}~\cite{Yi2014CASIA}      & {99.50} & {98.40} & {49.48} & {87.47} & {92.43} & {82.87} \\
{VGGFace2, R50, ArcFace}~\cite{cao2018vggface2}           & {99.60} & {98.80} & {55.37} & {91.77} & {93.72} & {85.02} \\ \midrule
{MS1MV2, R100, Arcface}~\cite{deng2019arcface}            & {99.77} & {99.65} & {64.48} & {92.50} & {95.83} & {90.13} \\
{MS1MV2, R100, Curricularface}~\cite{huang2020curricularface}     & {99.80} & {99.70} & {69.32} & {93.15} & {95.97} & {90.60} \\
{MS1MV2, R100, SFace}~\cite{zhong2021sface}              & {99.82} & {99.68} & {64.47} & {93.28} & {95.83} & {90.63} \\ 
\bottomrule
\end{tabular}
}
	\end{center}
	%\vspace{-0.1cm}
	\caption{Comparison of verification accuracy (\%) on MLFW database, as well as LFW~\cite{huang2008labeled}, SLLFW~\cite{deng2017fine}, TALFW~\cite{zhong2020towards}, CPLFW~\cite{zheng2018cross}, and CALFW~\cite{zheng2017cross} datbases using six high-performanced deep face recognition models.}
	\label{tab:table1}
\end{table*}

\begin{figure*}[t]
	\begin{center}
		\includegraphics[width=1\linewidth]{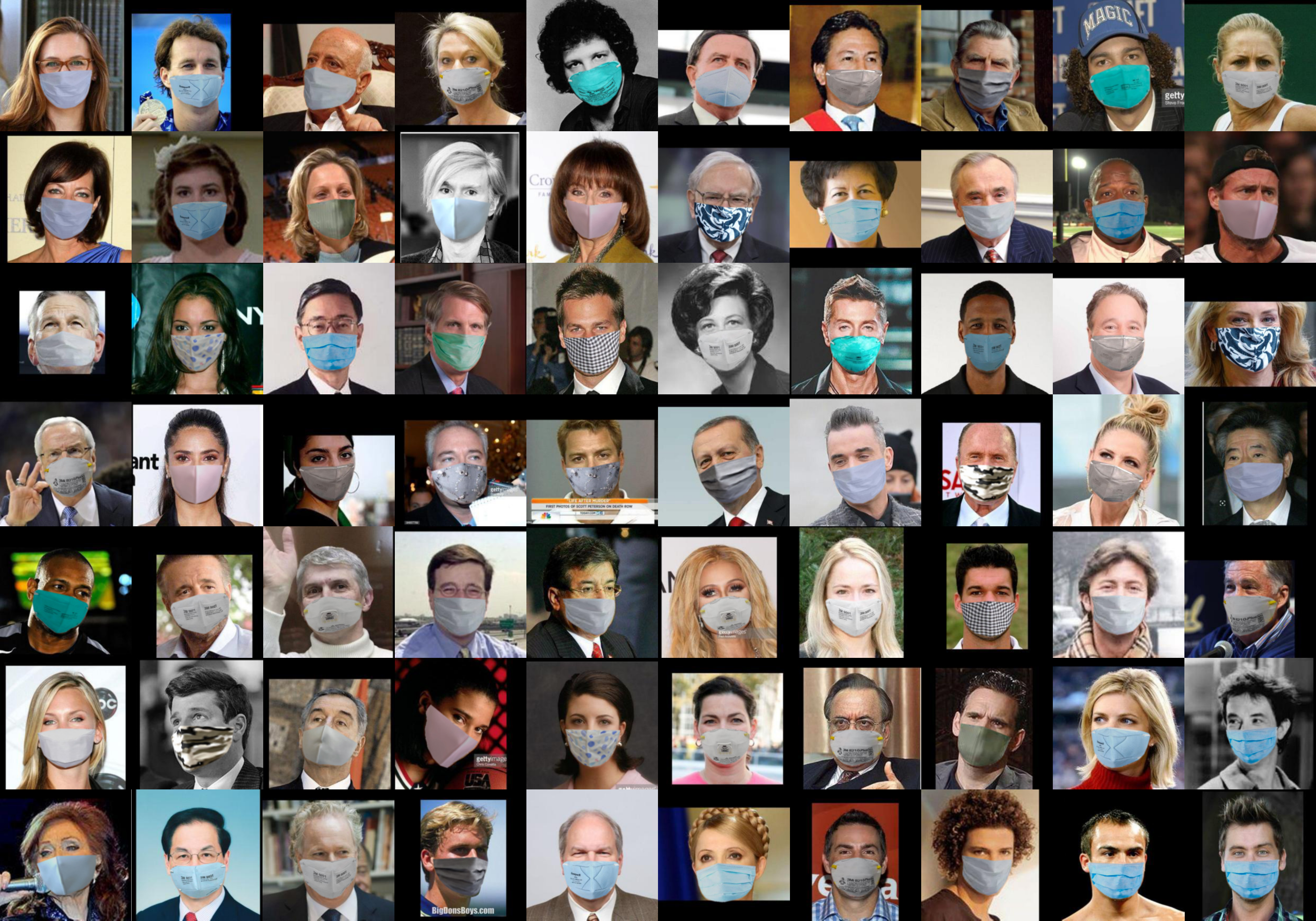}
	\end{center}
	\caption{
		Examples of masked faces generated by our mask tool.
	}
	\label{examples1}
\end{figure*}

\textbf{Generating masked faces.}
With the support of 31 mask templates, we generate the MLFW database according to the aforementioned settings.
For each face image in MLFW database, we provide the masked image of original size $250 \times 250$, and the landmarks used for alignment.
Pairing information is also stored and we use different suffixes, such as \textit{\_0001} and \textit{\_0002}, to distinguish the same original image of different masks.
The examples of masked faces generated by our tool are shown in Figure~\ref{examples1}.

\section{Baseline}
To evaluate the recognition performance on Masked LFW (MLFW) database, we select six opensourced SOTA deep face recognition methods as follows: (1) ResNet50 model trained on a private asia face dataset~\cite{wang2021face} with ArcFace~\cite{deng2019arcface}, (2) ResNet50 model trained on CASIA-WebFace database~\cite{Yi2014CASIA} with ArcFace~\cite{deng2019arcface}, (3) ResNet50 model trained on VGGFace2 database~\cite{cao2018vggface2} with ArcFace~\cite{deng2019arcface}, (4) ResNet100 model trained on MS1MV2 database~\cite{guo2016ms} refined by insightface with~\cite{deng2019arcface}, (5) ResNet100 model trained on MS1MV2 database~\cite{guo2016ms} with CurricularFace~\cite{huang2020curricularface}, (6) ResNet100 model trained on MS1MV2 database~\cite{guo2016ms} with SFace~\cite{zhong2021sface}.

In addition to the face verification performance (\%) on MLFW database, we also report accuracy (\%) on Labeled Faces in the Wild (LFW) database \cite{huang2008labeled}, Similar-looking LFW (SLLFW) database~\cite{deng2017fine}, Transferable Adversarial LFW database~\cite{zhong2020towards}, Cross-pose LFW (CPLFW) and Cross-age LFW (CALFW) for comprehensive evaluation.  

As shown in Table~\ref{tab:table1}, the accuracy of SOTA model on MLFW database is about 5\%-16\% lower than that on CALFW database, which demonstrates that SOTA methods also can not be directly used to recognize masked faces. 

\section{Conclusion}
In this paper, we aim to investigate the performance of face recognition models on masked faces. To this end, we have introduced a mask generation tool and built a test dataset, MLFW database. Finally, we have demonstrated that the recognition performance of face recognition models declines significantly on the masked face dataset.

{\small
	\bibliographystyle{ieee_fullname}
	\bibliography{egbib}

\begin{thebibliography}{10}\itemsep=-1pt

\bibitem{anwar2020masked}
Aqeel Anwar and Arijit Raychowdhury.
\newblock Masked face recognition for secure authentication.
\newblock {\em arXiv preprint arXiv:2008.11104}, 2020.

\bibitem{boutros2021mfr}
Fadi Boutros, Naser Damer, Jan~Niklas Kolf, Kiran Raja, Florian Kirchbuchner,
  Raghavendra Ramachandra, Arjan Kuijper, Pengcheng Fang, Chao Zhang, Fei Wang,
  et~al.
\newblock Mfr 2021: Masked face recognition competition.
\newblock In {\em 2021 IEEE International Joint Conference on Biometrics
  (IJCB)}, pages 1--10. IEEE, 2021.

\bibitem{bulat2017far}
Adrian Bulat and Georgios Tzimiropoulos.
\newblock How far are we from solving the 2d \& 3d face alignment problem? (and
  a dataset of 230,000 3d facial landmarks).
\newblock In {\em International Conference on Computer Vision}, 2017.

\bibitem{cabani2021maskedface}
Adnane Cabani, Karim Hammoudi, Halim Benhabiles, and Mahmoud Melkemi.
\newblock Maskedface-net--a dataset of correctly/incorrectly masked face images
  in the context of covid-19.
\newblock {\em Smart Health}, 19:100144, 2021.

\bibitem{cao2018vggface2}
Qiong Cao, Li Shen, Weidi Xie, Omkar~M Parkhi, and Andrew Zisserman.
\newblock Vggface2: A dataset for recognising faces across pose and age.
\newblock In {\em 2018 13th IEEE International Conference on Automatic Face \&
  Gesture Recognition (FG 2018)}, pages 67--74. IEEE, 2018.

\bibitem{deng2021masked}
Jiankang Deng, Jia Guo, Xiang An, Zheng Zhu, and Stefanos Zafeiriou.
\newblock Masked face recognition challenge: The insightface track report.
\newblock {\em arXiv preprint arXiv:2108.08191}, 2021.

\bibitem{deng2019arcface}
Jiankang Deng, Jia Guo, Niannan Xue, and Stefanos Zafeiriou.
\newblock Arcface: Additive angular margin loss for deep face recognition.
\newblock In {\em Proceedings of the IEEE Conference on Computer Vision and
  Pattern Recognition}, pages 4690--4699, 2019.

\bibitem{deng2017fine}
Weihong Deng, Jiani Hu, Nanhai Zhang, Binghui Chen, and Jun Guo.
\newblock Fine-grained face verification: Fglfw database, baselines, and
  human-dcmn partnership.
\newblock {\em Pattern Recognition}, 66:63--73, 2017.

\bibitem{du2021towards}
Hang Du, Hailin Shi, Yinglu Liu, Dan Zeng, and Tao Mei.
\newblock Towards nir-vis masked face recognition.
\newblock {\em IEEE Signal Processing Letters}, 28:768--772, 2021.

\bibitem{guo2016ms}
Yandong Guo, Lei Zhang, Yuxiao Hu, Xiaodong He, and Jianfeng Gao.
\newblock Ms-celeb-1m: A dataset and benchmark for large-scale face
  recognition.
\newblock In {\em European conference on computer vision}, pages 87--102.
  Springer, 2016.

\bibitem{huang2008labeled}
Gary~B Huang, Marwan Mattar, Tamara Berg, and Eric Learned-Miller.
\newblock Labeled faces in the wild: A database forstudying face recognition in
  unconstrained environments.
\newblock 2008.

\bibitem{huang2020curricularface}
Yuge Huang, Yuhan Wang, Ying Tai, Xiaoming Liu, Pengcheng Shen, Shaoxin Li,
  Jilin Li, and Feiyue Huang.
\newblock Curricularface: adaptive curriculum learning loss for deep face
  recognition.
\newblock In {\em Proceedings of the IEEE/CVF Conference on Computer Vision and
  Pattern Recognition}, pages 5901--5910, 2020.

\bibitem{wang2021face}
Qingzhong Wang, Pengfei Zhang, Haoyi Xiong, and Jian Zhao.
\newblock Face. evolve: A high-performance face recognition library.
\newblock {\em arXiv preprint arXiv:2107.08621}, 2021.

\bibitem{wang2020masked}
Zhongyuan Wang, Guangcheng Wang, Baojin Huang, Zhangyang Xiong, Qi Hong, Hao
  Wu, Peng Yi, Kui Jiang, Nanxi Wang, Yingjiao Pei, et~al.
\newblock Masked face recognition dataset and application.
\newblock {\em arXiv preprint arXiv:2003.09093}, 2020.

\bibitem{Yi2014CASIA}
Dong Yi, Zhen Lei, Shengcai Liao, and Stan~Z. Li.
\newblock Learning face representation from scratch.
\newblock {\em arXiv preprint arXiv:1411.7923}, 2014.

\bibitem{zheng2018cross}
Tianyue Zheng and Weihong Deng.
\newblock Cross-pose lfw: A database for studying cross-pose face recognition
  in unconstrained environments.
\newblock {\em Beijing University of Posts and Telecommunications, Tech. Rep},
  5, 2018.

\bibitem{zheng2017cross}
Tianyue Zheng, Weihong Deng, and Jiani Hu.
\newblock Cross-age lfw: A database for studying cross-age face recognition in
  unconstrained environments.
\newblock {\em arXiv preprint arXiv:1708.08197}, 2017.

\bibitem{zhong2020towards}
Yaoyao Zhong and Weihong Deng.
\newblock Towards transferable adversarial attack against deep face
  recognition.
\newblock {\em IEEE Transactions on Information Forensics and Security},
  16:1452--1466, 2020.

\bibitem{zhong2021sface}
Yaoyao Zhong, Weihong Deng, Jiani Hu, Dongyue Zhao, Xian Li, and Dongchao Wen.
\newblock Sface: sigmoid-constrained hypersphere loss for robust face
  recognition.
\newblock {\em IEEE Transactions on Image Processing}, 30:2587--2598, 2021.

\bibitem{zhu2021masked}
Zheng Zhu, Guan Huang, Jiankang Deng, Yun Ye, Junjie Huang, Xinze Chen, Jiagang
  Zhu, Tian Yang, Jia Guo, Jiwen Lu, et~al.
\newblock Masked face recognition challenge: The webface260m track report.
\newblock {\em arXiv preprint arXiv:2108.07189}, 2021.

\end{thebibliography}
}

\end{document}